\documentclass[11pt]{article}
\usepackage[margin=1.05in]{geometry}
\usepackage{amsmath,amssymb,amsthm}
\usepackage{bm}
\usepackage{booktabs}
\usepackage{graphicx}
\usepackage[hidelinks]{hyperref}
\usepackage{placeins}

\newcommand{\R}{\mathbb{R}}
\newcommand{\E}{\mathbb{E}}
\newcommand{\diag}{\operatorname{diag}}
\newcommand{\kraw}{\kappa_{\mathrm{raw}}}
\newcommand{\kadam}{\kappa_{\mathrm{adam}}}
\newcommand{\lmax}{\lambda_{\max}}
\newcommand{\lmin}{\lambda_{\min}}
\newcommand{\Ht}{\widetilde{H}}

\newtheorem{definition}{Definition}
\newtheorem{proposition}{Proposition}
\newtheorem{remark}{Remark}

\title{Loss Landscape Features That Make Adam Stall:\\
Definitions, Estimators, and the Preconditioned Hessian View}
\author{Rodion Podorozhny \\ Texas State University \\ rp31@txstate.edu }
\date{August 2026}

\begin{document}
\maketitle

\begin{abstract}
Across implicit-neural-representation (INR)  architectures and analytic
benchmarks we observe that a thoroughly tuned Adam (especially its learning rate (lr), e.g. in a hyperparameter sweep from $lr = 0.05$ to $10^{-8}$) can potentially reach a very low loss even on an ill-conditioned loss landscape or converge at a plateau far above the loss attained by second-order methods. 
This report defines the measured metrics that help determine if Adam can mitigate the ill-conditioning on a given loss landscape. It provides the indicators by which each outcome is determined, that are:
the condition number of the Hessian and of the
Adam-preconditioned Hessian $D^{-1/2}HD^{-1/2}$ 
(with the derivation from Adam's update rule), the diagonal mass $\rho$
that distinguishes axis-aligned from cross-coupled ill-conditioning,
the negative spectral mass estimated by stochastic Lanczos quadrature,
and the gradient energy fractions over curvature bands, including the
flat fraction that indicates the Adam stall. A worked out
$2\times 2$ example and an illustration show the reasons why a diagonal
preconditioning by Adam can remove axis-aligned ill-conditioning by rescaling and why it cannot do the same if the ill-conditioning is cross coupled. In addition, we present a case study of FINER image fitting architecture that goes over the whole loss landscape analysis
framework: the fitting architecture description, reasons due to which its landscape
stalls Adam at saddles, the measured PSNR values through our tuned baselines to the $120$--$134$\,dB results of the blockwise second order methods, the
error maps behind those numbers, and description of the benefits such image fitting accuracy gives in
practice.
\end{abstract}

\section{Experiment design and notation}

Let $f:\R^n\to\R$ be the training loss, $g(\theta)=\nabla f(\theta)$
its gradient and $H(\theta)=\nabla^2 f(\theta)$ its Hessian, with
eigendecomposition
\begin{equation}
H \;=\; \sum_{j=1}^{n} \lambda_j\, u_j u_j^\top,
\qquad \lambda_1 \ge \lambda_2 \ge \dots \ge \lambda_n,
\qquad u_j^\top u_i = \delta_{ij}.
\label{eq:eig}
\end{equation}
All image fitting experiments in this study are full-batch (a single image
or point set defines a deterministic loss), so $g$ and $H$ carry no
sampling noise; the estimators mentioned below randomize only the probe
vectors used for matrix-free computation of $H$.

\paragraph{Definition of "energy" term usage.}
Throughout, the \emph{energy} of a vector over an orthonormal system
denotes its squared $\ell_2$ magnitude resolved over that system: a
cumulative quantity that, like an area, adds over orthogonal
components, with Parseval's identity keeping the total invariant under
an orthonormal change of basis. For the gradient, the energy carried
by the eigendirection $u_j$ is $c_j^2$ with $c_j=u_j^\top g$, and
$\sum_j c_j^2 = \|g\|^2$; the partition of $\|g\|^2$ into fractions
over spectral bands is therefore well defined.

\paragraph{Matrix-free computation.}
Except for the small scale benchmarks (Rosenbrock, multi-saddle, 35k parameter SIREN), the full exact $H$ is not computed. Instead, for larger architectures (from around 200k parameters for FINER to 91.4 mln parameters for ViSIR), we use
Hessian--vector products (HVPs) by the Pearlmutter
construction~\cite{pearlmutter},
\begin{equation}
Hv \;=\; \nabla_\theta\bigl(g(\theta)^\top v\bigr).
\end{equation}

After computing and retaining the differentiable gradient
$g(\theta)=\nabla_\theta f(\theta)$, each additional HVP is obtained by
one reverse differentiation of $g(\theta)^\top v$.  Including the
construction of $g$, the cost is on the order of two gradient-equivalents,
with implementation dependent constant factors.

\section{Transformation of Hessian by Adam}
\label{sec:adam}

Adam~\cite{adam} computes the exponential moving averages
\begin{equation}
m_t = \beta_1 m_{t-1} + (1-\beta_1)\, g_t, \qquad
v_t = \beta_2 v_{t-1} + (1-\beta_2)\, g_t \odot g_t,
\end{equation}
with bias corrections $\hat m_t = m_t/(1-\beta_1^t)$,
$\hat v_t = v_t/(1-\beta_2^t)$, and updates
\begin{equation}
\theta_{t+1} \;=\; \theta_t \;-\; \alpha\,
\frac{\hat m_t}{\sqrt{\hat v_t} + \varepsilon}
\qquad\text{(all operations coordinatewise).}
\label{eq:adam}
\end{equation}
Let us define the diagonal matrix
\begin{equation}
D_t \;=\; \diag\!\bigl(\sqrt{\hat v_t} + \varepsilon\bigr)
\;\succ\; 0 .
\label{eq:D}
\end{equation}
Setting $\beta_1=0$ (momentum is discussed in
Remark~\ref{rem:momentum}), the update \eqref{eq:adam} is exactly
\begin{equation}
\theta_{t+1} \;=\; \theta_t \;-\; \alpha\, D_t^{-1} g_t :
\label{eq:precgd}
\end{equation}
which is gradient descent with the \emph{diagonal preconditioner} $D_t^{-1}$.
Because $\hat v_t$ is an exponential moving average with
$\beta_2 \in \{0.99,\dots,0.9999\}$, $D_t$ varies slowly relative to
the iteration; over the window in which convergence behavior is
evaluated we treat it as a constant $D$ (the quasi-static
approximation), for which we compute $\kadam$ (effective condition of Hessian transformed by Adam).

Two short derivations show that the curvature governing
\eqref{eq:precgd} is that of the symmetric transformed Hessian
\begin{equation}
\Ht \;=\; D^{-1/2} H\, D^{-1/2}.
\label{eq:Ht}
\end{equation}

\paragraph{Derivation 1: change of variables.}
Rescale coordinates by $\varphi = D^{1/2}\theta$ and define the
reparametrized loss $\tilde f(\varphi) = f(D^{-1/2}\varphi)$. By the
chain rule,
\begin{equation}
\nabla_\varphi \tilde f(\varphi) = D^{-1/2}\, g(\theta),
\qquad
\nabla^2_\varphi \tilde f(\varphi) = D^{-1/2} H(\theta)\, D^{-1/2}
= \Ht .
\end{equation}
Plain gradient descent on $\tilde f$,
$\varphi_{t+1} = \varphi_t - \alpha \nabla_\varphi\tilde
f(\varphi_t)$, maps back through $\theta = D^{-1/2}\varphi$ to
\begin{equation}
D^{1/2}\theta_{t+1} = D^{1/2}\theta_t - \alpha D^{-1/2} g_t
\quad\Longleftrightarrow\quad
\theta_{t+1} = \theta_t - \alpha D^{-1} g_t,
\end{equation}
which is \eqref{eq:precgd}. Momentum-less Adam restated via quasi-static $D$ is
therefore \emph{plain gradient descent on the reparametrized loss},
and the Hessian of that loss, the curvature the method actually
encounters, is $\Ht$. Division of each gradient coordinate by
$\sqrt{\hat v_i}+\varepsilon$ in \eqref{eq:adam} is precisely the
$D^{-1}$ in \eqref{eq:precgd}, and the symmetric split of $D^{-1}$
around $H$ in \eqref{eq:Ht} is what that division looks like after
the change of variables that turns an iteration into a plain gradient
step.

\paragraph{Derivation 2: the iteration matrix.}
On a local quadratic model
$f(\theta) = f(\theta^\star) + \tfrac12 (\theta-\theta^\star)^\top H
(\theta-\theta^\star)$ with error $e_t = \theta_t - \theta^\star$,
update \eqref{eq:precgd} gives $g_t = H e_t$ and
\begin{equation}
e_{t+1} \;=\; \bigl(I - \alpha D^{-1} H\bigr)\, e_t .
\end{equation}
The iteration matrix is similar to a symmetric one:
\begin{equation}
D^{1/2}\bigl(I - \alpha D^{-1}H\bigr)D^{-1/2}
\;=\; I - \alpha\, D^{-1/2} H D^{-1/2}
\;=\; I - \alpha \Ht ,
\end{equation}
so $\operatorname{spec}(I-\alpha D^{-1}H) =
\operatorname{spec}(I-\alpha\Ht)$. Stability requires
$\alpha < 2/\lmax(\Ht)$, and with the optimal step
$\alpha^\star = 2/(\lmax(\Ht)+\lmin(\Ht))$ the asymptotic error
contraction per iteration is
\begin{equation}
r^\star \;=\; \frac{\kappa(\Ht) - 1}{\kappa(\Ht) + 1},
\qquad
\kappa(\Ht) = \frac{\lmax(\Ht)}{\lmin(\Ht)} ,
\label{eq:rate}
\end{equation}
it is the first-order optimization contraction rate with $\kappa(\Ht)$ in place of
$\kappa(H)$. Both derivations show that the
condition number on which Adam operates is that of $\Ht$.

\begin{remark}[Momentum and $\varepsilon$]\label{rem:momentum}
Momentum ($\beta_1>0$) can accelerate the preconditioned iteration in the
manner of the heavy-ball method, but it does not alter the local geometry:
the iteration is still governed by the same rescaled Hessian $\Ht$.  For a
positive-definite quadratic with optimally tuned momentum, the iteration
complexity for a fixed multiplicative reduction in error improves from
$O(\kappa)$ to $O(\sqrt{\kappa})$ at best; momentum does not itself reduce
$\kappa$.

On the benchmarks and architectures in this study on which Adam stalls,
the measured $\kadam$ is of order $10^5$ to $10^6$.  Thus, even the ideal
accelerated dependence remains $\sqrt{\kadam}\approx
3\times10^2$ to $10^3$, so momentum can mitigate but cannot eliminate the
conditioning barrier left after diagonal rescaling.

The floor $\varepsilon$ ensures $D\succeq\varepsilon I$ and prevents
division by near-zero coordinate scales when $\hat v_i\to0$.  Along the evaluated trajectories, $\varepsilon$ is dominated by
$\sqrt{\hat v_i}$ except in the flattest coordinates.
\end{remark}

\begin{definition}[Raw and Adam-preconditioned condition numbers]
\label{def:kappa}
At a trajectory checkpoint $\theta_t$ with Adam state $\hat v_t$,
\begin{equation}
\kraw \;=\; \frac{\lmax(H)}{|\lmin(H)|},
\qquad
\kadam \;=\; \frac{\lmax(\Ht)}{|\lmin(\Ht)|},
\qquad
\Ht = D_t^{-1/2} H(\theta_t)\, D_t^{-1/2},
\end{equation}
where the extreme eigenvalues are obtained from $m$-step Lanczos with
full reorthogonalization ($m=96$) on the HVP oracle; for $\Ht$ the
oracle is $u \mapsto D^{-1/2}\,H\,(D^{-1/2}u)$, two diagonal scalings
around one HVP. The \emph{reduction factor} is $\kraw/\kadam$. When
$\lmin<0$ with $|\lmin|$ comparable to $\lmax$ the ratio is reported
as indefinite (the multi-saddle and Chebyshev--Rosenbrock rows of
Table~\ref{tab:fingerprint}, discussed in Section~\ref{sec:table}).
\end{definition}

\section{Axis-aligned versus cross-coupled ill-conditioning}
\label{sec:coupling}

A diagonal preconditioner rescales coordinate axes; it cannot rotate
them. Whether $\kadam \ll \kraw$ is therefore a question about where
the ill-conditioning of $H$ is located relative to the coordinate basis.

\begin{definition}[Diagonal mass]\label{def:rho}
\begin{equation}
\rho(H) \;=\;
\frac{\|\diag(H)\|_F^2}{\|H\|_F^2}
\;=\;
\frac{\sum_i H_{ii}^2}{\sum_{i,j} H_{ij}^2}
\;\in\; (0,1].
\end{equation}
$\rho = 1$ iff $H$ is diagonal (eigenbasis equal to the coordinate
basis). Small $\rho$ means the Frobenius mass is carried by
off-diagonal entries: stiff and soft eigendirections are mixtures of
coordinates. We call the ill-conditioning \emph{axis-aligned} when
$\rho$ is large and \emph{cross-coupled} when $\rho$ is small.
\end{definition}

\subsection{A worked $2\times 2$ example}
\label{sec:2x2}

Fix $L \gg \mu > 0$ and compare
\begin{equation}
H_1 = \begin{pmatrix} L & 0\\ 0 & \mu \end{pmatrix},
\qquad
H_2 = R_{45^\circ}^\top
\begin{pmatrix} L & 0\\ 0 & \mu \end{pmatrix}
R_{45^\circ}
= \frac12
\begin{pmatrix}
L+\mu & L-\mu\\
L-\mu & L+\mu
\end{pmatrix},
\end{equation}
the same spectrum $\{L, \mu\}$, $\kappa(H_1)=\kappa(H_2)=L/\mu$, once
with the stiff direction along a coordinate axis and once rotated by
$45^\circ$.

\emph{Axis-aligned case.} Scaling $H_1$ by its own diagonal,
$D = \diag(H_1)$, gives
$D^{-1/2} H_1 D^{-1/2} = I$: condition number exactly $1$. A diagonal
preconditioner significantly reduces  axis-aligned ill-conditioning.
This is the mechanism behind the condition reduction factors of
$16$ times to $887$ times on the INR architectures, where $\rho$ is large or
the stiff directions align with coordinates. In such cases Adam performs as well as second order methods,
provided Adam does not come across saddles on its search trajectory.

\emph{Cross-coupled case.} $\diag(H_2) = \tfrac{L+\mu}{2} I$ is a
multiple of the identity, so scaling by it (indeed, by \emph{any}
diagonal $D = \delta I$ that respects the symmetry between the two
coordinates) leaves the condition number untouched:
\begin{equation}
\kappa\!\left(D^{-1/2} H_2 D^{-1/2}\right) = \kappa(H_2) = \frac{L}{\mu}
\qquad \text{for every } D = \delta I .
\end{equation}
More generally, minimizing over all positive diagonals cannot reach
below $\kappa(H_2)/\!\left(\text{small factor}\right)$ here, because
the stiff direction $\tfrac{1}{\sqrt2}(1,1)^\top$ and the soft
direction $\tfrac{1}{\sqrt2}(1,-1)^\top$ load both coordinates
equally: no per-coordinate rescaling can tell them apart. The
reduction factor is $\approx 1$, and Adam faces the same $L/\mu$ that
plain gradient descent faces. For this example
$\rho(H_2) = \tfrac{(L+\mu)^2}{(L+\mu)^2 + (L-\mu)^2} \to \tfrac12$ as
$L/\mu \to \infty$; in high dimension with dense coupling $\rho$
falls much lower (the measured coupled benchmarks have
$\rho \le 0.04$). 

\begin{figure}[!htbp]
\centering
\includegraphics[width=0.86\textwidth]{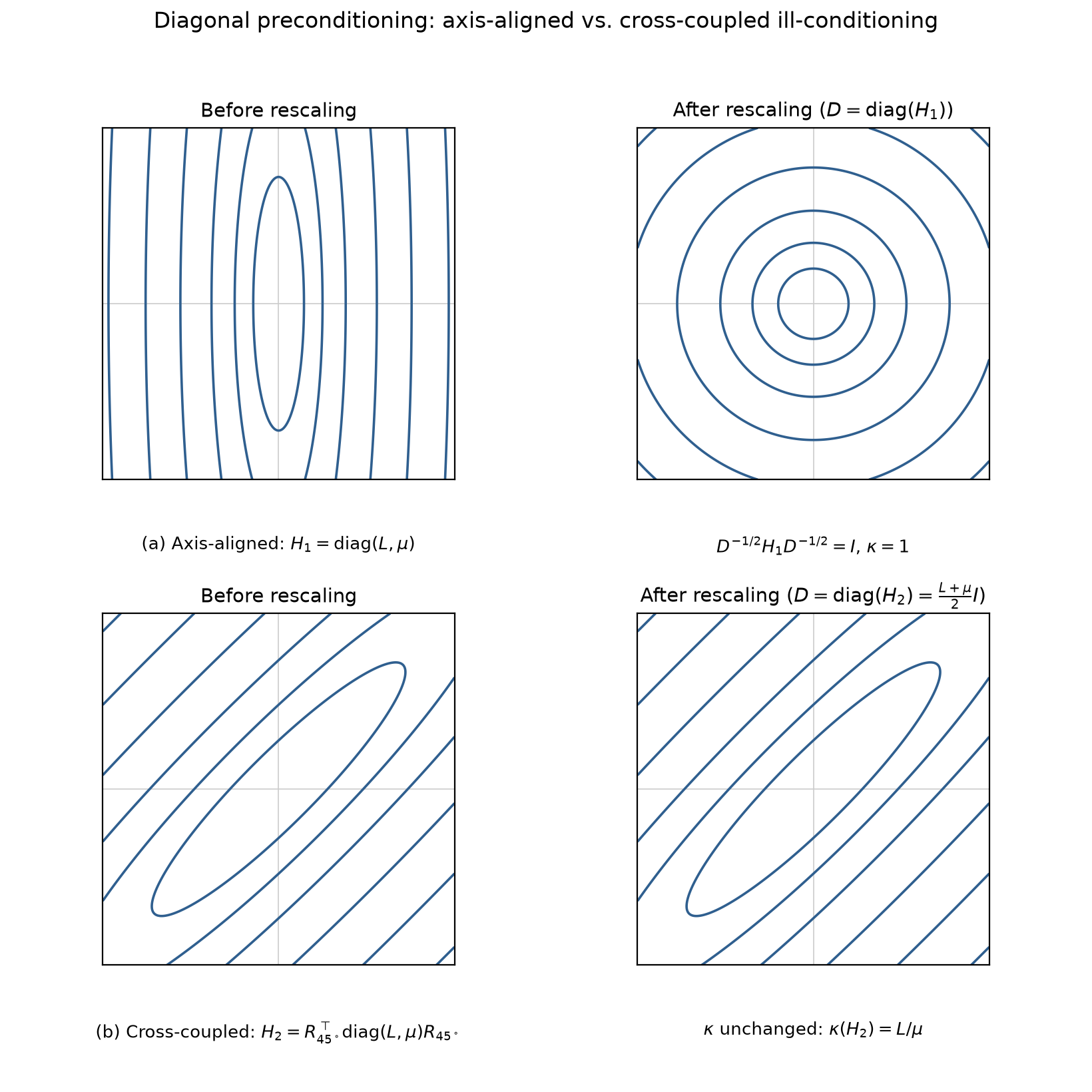}
\caption{Diagonal preconditioning on the two $2\times2$ examples of
Section~\ref{sec:2x2}. Contours of the local quadratic model before
(left) and after (right) the rescaling
$\varphi = D^{1/2}\theta$ with $D = \diag(H)$. (a) When the stiff
direction is a coordinate axis, per-coordinate rescaling equalizes the
curvature scales and the contours become circles: Adam behaves as if
the problem were well conditioned. (b) When stiff and soft directions
are mixtures of the coordinates, the diagonal of $H$ is uniform, the
rescaling is a multiple of the identity, and the contours, and
$\kappa$, are unchanged: Adam faces the raw ill-conditioning, and
by van der Sluis' theorem no other diagonal would do better than a
modest factor.}
\label{fig:coupling}
\end{figure}

An illustration of axis-aligned versus cross-coupled case is given in Fig.~\ref{fig:coupling}

\FloatBarrier

\subsection{A diagonal-scaling certificate: van der Sluis' theorem}

The measured $\kadam$ uses Adam's particular diagonal scaling $D_t$.
A large value of $\kadam$ therefore shows that Adam's
gradient statistics based scaling has not removed the conditioning
barrier. To determine if this barrier is specific to \emph{all}
per-coordinate rescalings, consider the following result.

\begin{proposition}[van der Sluis~\cite{vandersluis}]
Let $A\in\R^{n\times n}$ be symmetric positive definite, and let
$D_A=\diag(A)$. Then
\begin{equation}
\kappa\!\left(D_A^{-1/2} A D_A^{-1/2}\right)
\;\le\;
m\,
\min_{\substack{D\succ0\\D\ \mathrm{diagonal}}}
\kappa\!\left(D^{-1/2} A D^{-1/2}\right),
\end{equation}
where $m\le n$ is the maximum number of nonzero entries in a row of
$A$.
\end{proposition}

Thus, Jacobi scaling by the curvature diagonal is within a factor $m$
of the best symmetric positive diagonal scaling. Equivalently,
\begin{equation}
\min_{\substack{D\succ0\\D\ \mathrm{diagonal}}}
\kappa\!\left(D^{-1/2} A D^{-1/2}\right)
\;\ge\;
\frac{1}{m}\,
\kappa\!\left(D_A^{-1/2} A D_A^{-1/2}\right).
\label{eq:vandersluis-lower}
\end{equation}
Consequently, if the Jacobi scaled curvature remains large relative to
$m$, no positive per-coordinate rescaling can make the local quadratic
well conditioned.

Adam's scale $\sqrt{\hat v_t}$ is derived from coordinatewise
gradient statistics rather than from $\diag(A)$. It may serve as a
heuristic surrogate for coordinatewise curvature, but van der Sluis'
theorem does not imply that it approximates $D_A$. Accordingly, a large
$\kadam$ diagnoses the failure of Adam's particular diagonal scaling;
the stronger conclusion in \eqref{eq:vandersluis-lower} requires
measuring the Jacobi scaled condition number.

When this stronger condition holds, progress along coupled directions
requires a method that represents cross coordinate curvature, such as
block preconditioning, low-rank curvature models, or Krylov based
cubic regularized steps.

\section{Matrix-free estimation of the spectral features }
\label{sec:estimators}

\subsection{Extreme eigenvalues}
$\lmax$ and $\lmin$ are the extreme Ritz values of $m$-step Lanczos
with full reorthogonalization ($m = 96$) on the corresponding oracle
($Hv$ for $\kraw$; $D^{-1/2}HD^{-1/2}v$ for $\kadam$). Ritz values
converge to the spectrum edges first, so moderate $m$ suffices for the
extreme eigenvalues even when $n \sim 10^5$--$10^8$.

\subsection{Negative spectral mass by stochastic Lanczos quadrature}
\label{sec:slq}

\begin{definition}[Spectral density and negative mass]
The (normalized) Hessian spectral density is the measure
\begin{equation}
\mu(\lambda) \;=\; \frac1n \sum_{j=1}^{n} \delta(\lambda - \lambda_j),
\end{equation}
and the negative spectral mass at relative threshold
$\epsilon = 10^{-3}$ is
\begin{equation}
m_{\mathrm{neg}} \;=\;
\int_{-\infty}^{-\epsilon\,\lmax} d\mu(\lambda)
\;=\;
\frac{\#\{\, j : \lambda_j < -\epsilon\, \lmax \,\}}{n}.
\label{eq:negmass}
\end{equation}
\end{definition}

$m_{\mathrm{neg}}$ is estimated by stochastic Lanczos quadrature
(SLQ)~\cite{slq} in the following way.

For each stochastic probe, draw a vector $r\in\R^n$ with independent
Rademacher entries,
\begin{equation}
\Pr(r_i=+1)=\Pr(r_i=-1)=\tfrac12,
\qquad i=1,\ldots,n,
\end{equation}
and normalize it as
\begin{equation}
z=\frac{r}{\|r\|_2}=\frac{1}{\sqrt n}r.
\end{equation}
The latter equality holds exactly because $r_i^2=1$ for every $i$, and
therefore $\|r\|_2^2=n$. Thus $\|z\|_2=1$ and
$\E_z[zz^\top]=I/n$.

Let $\{(\lambda_j,u_j)\}_{j=1}^n$ be the eigenpairs of the symmetric
Hessian $H$, with $\{u_j\}_{j=1}^n$ an orthonormal eigenbasis.
Expanding the probe in this basis,
\begin{equation}
z=\sum_{j=1}^n (u_j^\top z)\,u_j,
\end{equation}
shows that $(u_j^\top z)^2$ is the fraction of the probe's unit
squared norm aligned with eigendirection $u_j$. The corresponding
probe-weighted spectral measure is
\begin{equation}
\mu_z(\lambda)
=
\sum_{j=1}^n (u_j^\top z)^2\,
\delta(\lambda-\lambda_j).
\label{eq:probe-spectral-measure}
\end{equation}

The squared coefficients $(u_j^\top z)^2$ define a probability weight
over the Hessian eigendirections, since
\begin{equation}
\sum_{j=1}^n (u_j^\top z)^2
=
\|z\|_2^2
=
1.
\end{equation}

Because $\E_z[zz^\top]=I/n$, every eigendirection receives equal
expected weight:
\begin{equation}
\E_z\!\left[(u_j^\top z)^2\right]
=
u_j^\top \E_z[zz^\top]u_j
=
\frac1n.
\end{equation}
Consequently,
\begin{equation}
\E_z\,\mu_z(\lambda)
=
\frac1n\sum_{j=1}^n\delta(\lambda-\lambda_j)
=
\mu(\lambda),
\end{equation}
where $\mu$ is the normalized empirical spectral measure of $H$.

Starting the Lanczos process with $q_1=z$ constructs the orthogonal
polynomials associated with the probe-weighted measure $\mu_z$. Its
Ritz values and quadrature weights therefore provide a Gauss--quadrature
approximation to spectral quantities under $\mu_z$. Averaging the
estimates over independent probes estimates the corresponding quantity
under the normalized empirical spectral measure $\mu$.
\iffalse
For a probe vector $z\in\R^n$ obtained by drawing independent
Rademacher entries $r_i\in\{-1,+1\}$ with equal probability and
normalizing,
expand $z = \sum_j (u_j^\top z)\, u_j$; the probe-weighted spectral
measure is then 
\begin{equation}
\mu_z(\lambda) = \sum_j (u_j^\top z)^2\, \delta(\lambda-\lambda_j),
\qquad
\E_z\,\mu_z = \mu
\end{equation}
is exactly the measure whose orthogonal polynomials are built by Lanczos process.
\fi

Performing $m$ Lanczos steps from $q_1 = z$ yields the tridiagonal
$T_m = Q_m^\top H Q_m$, and its eigendecomposition
$T_m = S\,\Theta\,S^\top$ delivers a Gauss quadrature rule for
$\mu_z$ with nodes and weights
\begin{equation}
\theta_i \;=\; \Theta_{ii},
\qquad
w_i \;=\; (S_{1i})^2 ,
\qquad
\int h \, d\mu_z \;\approx\; \sum_{i=1}^m w_i\, h(\theta_i).
\label{eq:quad}
\end{equation}

Averaging over $P$ probes and applying \eqref{eq:quad} to the
indicator of $(-\infty, -\epsilon\lmax)$,
\begin{equation}
\widehat m_{\mathrm{neg}}
\;=\;
\frac1P \sum_{p=1}^{P} \;\sum_{i \,:\, \theta_i^{(p)} <\,
-\epsilon\widehat\lmax} w_i^{(p)} .
\end{equation}

Here $\widehat\lmax$ is the Lanczos estimate of $\lmax$; hence
$\widehat m_{\mathrm{neg}}$ includes both quadrature error and threshold
error from estimating the spectral scale.

\paragraph{Reason for the relative threshold.}
The cutoff $-\epsilon\lmax$ (and the matching flat band below) is
proportional to $\lmax$ for two reasons. Scale invariance: replacing
$f$ by $c f$ multiplies every eigenvalue by $c$, so any absolute
cutoff would classify the same geometry differently under a trivial
rescaling of the loss. Dead-band: overparametrized networks carry a
large cluster of eigenvalues numerically indistinguishable from zero, the bulk of the spectrum, with only a handful of
data-determined outliers away from it~\cite{sagun}, and the
SLQ estimates resolve that cluster with finite precision. The
dead-band keeps the near-zero bulk, and the estimator noise, out of
both the flat and the saddle counts. Then the condition ``$\leq -10^{-3}\lmax$''
is interpreted as: negative with magnitude exceeding $0.1\%$ of the top
curvature scale, meaningfully negative at the problem's own scale.

\subsection{Gradient energy fractions over curvature bands}
\label{sec:gradenergy}

\begin{definition}[Band energies of the gradient]
With $c_j = u_j^\top g$, the fraction of gradient energy in a spectral
band $B \subset \R$ is
\begin{equation}
w(B) \;=\; \frac{1}{\|g\|^2} \sum_{j \,:\, \lambda_j \in B} c_j^2 .
\end{equation}
The reported bands are
\begin{equation}
\underbrace{\;\theta > 0.1\,\lmax\;}_{\textrm{stiff}}
\qquad
\underbrace{\;|\theta| \le 10^{-3}\lmax\;}_{\textrm{flat
(dead-band)}}
\qquad
\underbrace{\;\theta < -10^{-3}\lmax\;}_{\textrm{negative}},
\label{eq:bands}
\end{equation}
and $\mathrm{flat\ frac} = w\bigl(\{|\theta| \le
10^{-3}\lmax\}\bigr)$.
\end{definition}

No eigendecomposition is needed: running Lanczos \emph{from the
normalized gradient} $q_1 = g/\|g\|$ makes the quadrature weights
\eqref{eq:quad} approximate exactly the $g$-weighted spectral
distribution,
\begin{equation}
\mu_g(\lambda) = \frac{1}{\|g\|^2}\sum_j c_j^2\,
\delta(\lambda - \lambda_j)
\;\approx\;
\sum_{i=1}^{m} w_i \,\delta(\lambda - \theta_i),
\qquad
\widehat w(B) = \sum_{\theta_i \in B} w_i ,
\end{equation}
with no probe averaging (the starting vector is deterministic).

Three interpretations help understanding of these measured gradient
fractions. First, the bands \eqref{eq:bands} do not partition the
spectrum. The two decades of moderate curvature between
$10^{-3}\lmax$ and $0.1\lmax$ are deliberately unlabeled since they are neither effectively flat, strongly stiff, nor negative and therefore do not directly bear on the diagnostic’s purpose. 
On all the evaluated benchmarks the middle band carries nearly all of $\|g\|^2$. A flat fraction of 
0.000 means that none of the gradient energy lies in directions whose curvature is effectively zero at the scale of the problem. Second, the values are
late trajectory snapshots: a first order step reduces the gradient
component along $u_j$ at a rate proportional to $\lambda_j$ (on the
quadratic model, $c_j \mapsto (1-\alpha\lambda_j)c_j$; the same
rate rule underlies neural tangent kernel training dynamics
\cite{jacot}), so dead-band
components barely move while stiff ones reduce (are suppressed) quickly; transiently
large mid-run values in the \emph{flat} fraction (e.g.\ $.111$ on
SIREN~2D and $.689$ on SDF-SIREN, meaning $11.1\%$ and $69\%$ of
gradient energy transiently sit in the dead band, not the negative
spectral mass column of Table~\ref{tab:fingerprint})
are the spectral-bias stall phases, and their return to zero at the
endpoint means the tuned Adam optimizer run eventually progressed on these components.
Third, the key diagnostic is a \emph{persistently} nonzero flat fraction together with substantial negative-curvature mass. When training ends, still appreciable gradient energy remains in directions where first-order updates have little effect.

\subsection{Diagonal mass: exact and debiased probe estimators}
\label{sec:rhoest}

Where full exact $H$ is feasible to compute (e.g. the $35$k parameter SIREN benchmark), $\rho$ is
computed exactly from Definition~\ref{def:rho}. At scale it is
estimated with Rademacher probes $z_k \in \{\pm1\}^n$,
$k = 1,\dots,K$:
\begin{equation}
\widehat d \;=\; \frac1K \sum_{k} z_k \odot (H z_k),
\qquad
\E\,\widehat d = \diag(H),
\qquad
\widehat F \;=\; \frac1K \sum_{k} \|H z_k\|^2,
\qquad
\E\,\widehat F = \|H\|_F^2 .
\end{equation}
The naive plug-in $\|\widehat d\|^2$ overestimates
$\sum_i H_{ii}^2$ because each coordinate of a single probe carries
off-diagonal leakage,
$\E\bigl[(z\odot Hz)_i^2\bigr] = H_{ii}^2 + \sum_{j\ne i} H_{ij}^2$,
so with $K$ probes
\begin{equation}
\E\,\|\widehat d\|^2
= \sum_i H_{ii}^2 + \frac1K \sum_{i\ne j} H_{ij}^2
= S_{\mathrm d} + \frac{F - S_{\mathrm d}}{K},
\qquad
S_{\mathrm d} := \textstyle\sum_i H_{ii}^2,\;\; F := \|H\|_F^2 .
\end{equation}
Solving for $S_{\mathrm d}$ gives the debiased estimator used by the
diagonal-dominance router:
\begin{equation}
\widehat S_{\mathrm d}
= \frac{K \|\widehat d\|^2 - \widehat F}{K - 1},
\qquad
\widehat\rho
= \operatorname{clip}_{[0,1]}
\left(\widehat S_{\mathrm d} \,/\, \widehat F\right).
\label{eq:rhodebias}
\end{equation}

\section{The measured fingerprints}
\label{sec:table}

Table~\ref{tab:fingerprint} collects the features along the tuned Adam
trajectory (late-trajectory checkpoints; initialization or mid-run
values in the table notes where diagnostic). Learning rates and
$(\beta_2, \varepsilon)$ were set according to the published INR papers, so
the values in Adam column correspond to a tuned version (the best PSNR obtained from a hyperparemeter sweep of learning rate (lr) from $0.05$ to $10^{-8}$, 
which is in this case $lr = 10^{-4}$).

\begin{table}[t]
\centering
\footnotesize
\setlength{\tabcolsep}{4pt}
\begin{tabular}{lccccccl}
\toprule
Benchmark & $\kraw$ & $\kadam$ & reduction & $\rho$ & neg.\ mass &
flat frac & outcome \\
\midrule
35k multi-freq.\ & 4.9e2 & 3.1e1 & 16$\times$ & .04 & .92 & .000 &
Adam at precision floor \\
SIREN 2D & 1.6e3 & 9.9e1 & 16$\times$ & .09 & .09 & .000\textsuperscript{a} & comparable \\
WIRE 2D & 1.2e5 & 8.4e2 & 143$\times$ & .65 & .00 & .000 &
comparable \\
FINER 2D & 5.4e4 & 6.1e1 & 887$\times$ & .98 & .19\textsuperscript{b} & .000
& ARC $+59.5$ dB \\
SDF-SIREN & 8.9e3 & 2.4e1 & 372$\times$ & .06 & .67 & .000\textsuperscript{c} & comparable \\
SDF-FINER & 1.0e3 & 6.9 & 146$\times$ & .58 & .68 & .001 & Adam error
5.5$\times$ \\
multisaddle & 2.0e6 & 1.5e6 & 1.3$\times$ & .02 & .46 & .005 &
second order better \\
Chebyshev-Rosenbrock & indef. & 5.5e5 & --- & .04 & .49 & .001 & second order
better \\
\bottomrule
\end{tabular}
\caption{Landscape fingerprints along the tuned Adam trajectory.
Reduction $= \kraw/\kadam$. Architectures: SIREN~\cite{siren}, WIRE
\cite{wire}, FINER~\cite{liu2024finer}; the 35k multi-frequency
benchmark is a SIREN-family network with six octave-spaced target
frequencies~\cite{visir}; the SDF rows fit signed distance functions
(Thai statue) under IGR supervision~\cite{gropp} with SIREN and FINER
backbones. The two analytic rows are decided by final loss (the
benchmarks have no PSNR): the first-order arms terminate on the saddle
plateau at any tested tuning while cubic-regularized arms pass it by
orders of magnitude.
Notes: \textsuperscript{a}$.111$ mid-run;
\textsuperscript{b}$.81$ at initialization;
\textsuperscript{c}$.689$ transient mid-run.}
\label{tab:fingerprint}
\end{table}

The results fall into three categories.

\paragraph{Axis-aligned (diagonal-repairable).}
Large reduction factors ($16\times$--$887\times$): Adam removes the
raw ill-conditioning by rescaling the Hessian, exactly as in
Figure~\ref{fig:coupling}(a), and, in cases where negative mass is also benign
along the trajectory, the outcomes are comparable between tuned Adam
and the block second-order experiment arms (SIREN~2D, WIRE~2D, SDF-SIREN at the
achieved accuracy; the 35k SIREN benchmark reaches the machine-precision
floor).

\paragraph{Cross-coupled (the stall by conditioning).}
Multi-saddle and Chebyshev-Rosenbrock benchmarks keep $\kadam \ge 5.5\cdot10^5$ at reduction $\le 1.3\times$ with $\rho \le .04$: the ill-conditioning is due to
coupled directions, Figure~\ref{fig:coupling}(b), no diagonal can
reach it (Section~\ref{sec:coupling}), and by \eqref{eq:rate} the
per-iteration contraction is $1 - O(1/\kadam)$: about $10^{5}$--$10^{6}$
iterations per error along the soft coupled directions. That is, this results in 
a stall of a first order method at any practical budget, 
and, at larger scale,  it also appears on the 91.4M
ViSIR benchmark (diagonal ESGD 4.9~dB against 64.4~dB for the blockwise
cubic method).

\begin{figure}[!htbp]
\centering
\includegraphics[width=\textwidth]{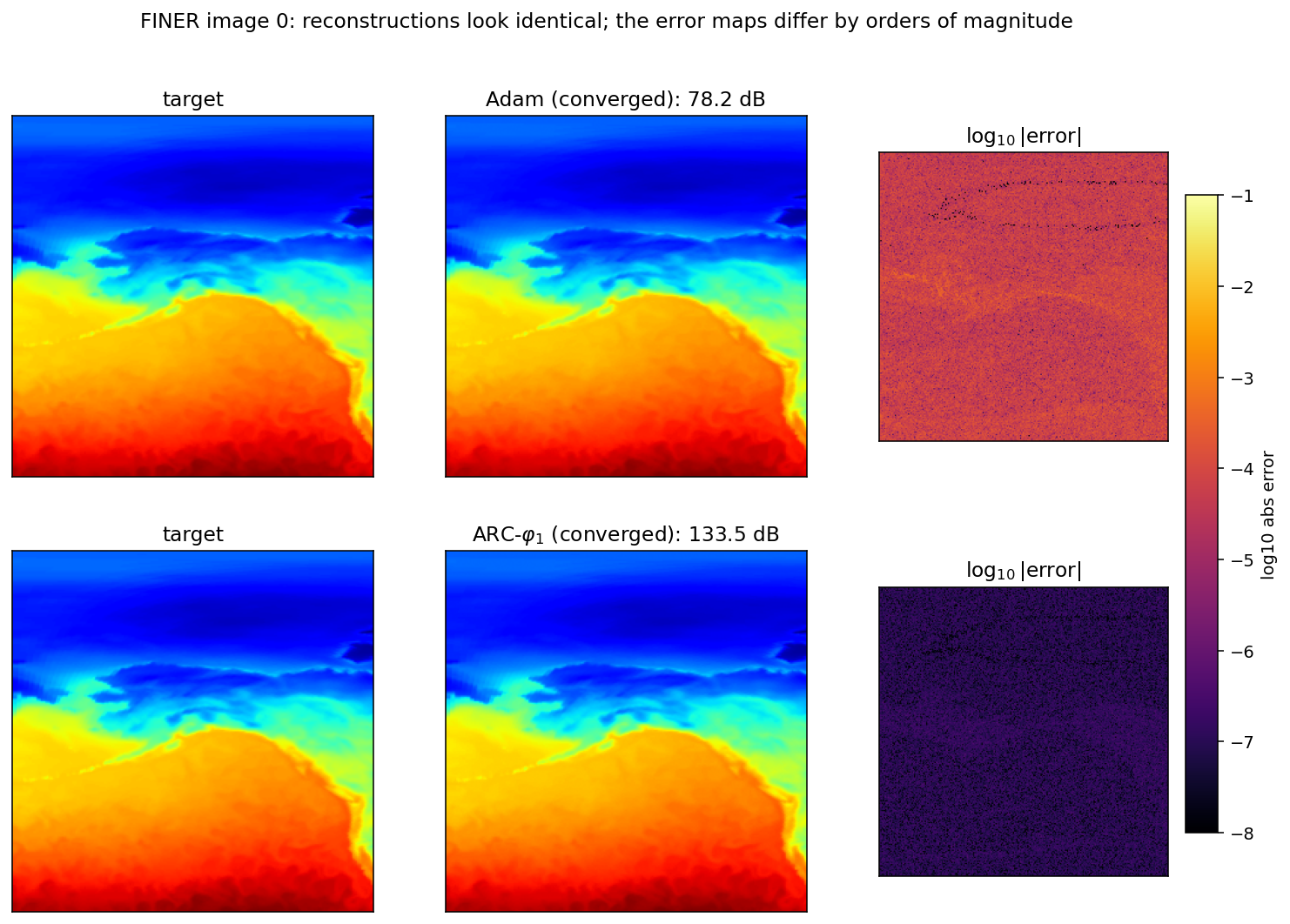}
\caption{FINER image 0 at native $240^2$ resolution: target,
reconstruction, and $\log_{10}$ absolute-error map for
extended-budget converged Adam ($78.2$~dB at $1{,}000{,}001$
iterations, top) and converged ARC-$\varphi_1$ ($133.5$~dB at
$2{,}824$ sweeps, bottom). The error-map color scale is shared
across both rows and spans $10^{-1}$ to $10^{-8}$. The
reconstructions are visually indistinguishable from the target and
from each other, both are far above the $\sim$50~dB perceptual
threshold. The error maps differ by nearly three orders of
magnitude and in \emph{structure}: Adam's residual is a smooth
$\sim$$10^{-4}$ field organized in broad regions separated by dark
zero-crossing curves, a low-frequency bias field, while the
ARC-$\varphi_1$ residual is a band-uniform $\sim$$2\cdot10^{-7}$
speckle with no spatial organization.}
\label{fig:native}
\end{figure}

\paragraph{Saddle-dominated (the stall at an Adam stationary point).}
FINER's variable-periodic activations put $0.81$ (out of 1.0) of the spectral mass
below $-10^{-3}\lmax$ at initialization. Tuned Adam converges (reaching its plateau), to a point where its own search step update norm vanishes while
the Hessian remains indefinite. Thus Adam reaches its \emph{stationary point} at
a saddle, at 65~dB with the residual concentrated at low
frequency. The loss is full-batch and deterministic: there is no
gradient noise to drive escape, and a first-order step has no
mechanism to exploit a direction in which the gradient component is
zero but the curvature is negative. A cubic regularized block step
does escape such a saddle: the shift $\lambda = M\|s\|/2 > -\lmin$ makes every subproblem
positive definite, so the step descends \emph{along} negative
curvature deterministically, reaching 120--129~dB with a band-uniform
residual. The same structure appears in 3D (SDF-FINER, negative mass
$.68$, Adam's reconstruction error $5.5\times$ that of the
second-order arm). To be sure that Adam reached its plateau, the hyperparameters sweep was conducted for 1 mln iterations for each value of learning rate on image 0 of Earth System Model. Adam's endpoint verified by convergence to a plateau on this image reaches $67.7$~dB at iteration $100{,}001$ ($421.5$~s, $\approx$7~min) on RTX~PRO~6000; our second order optimizer ARC-$\phi_1$~\cite{podorozhny2026arcphi1} (blockwise adaptive cubic regularized with a subproblem solver via an exponential relaxation function recurrence) reaches $125.5$~dB at $8{,}547$ sweeps in $7{,}201$~s ($\approx$120~min). Extending Adam's own budget to a flat $1{,}000{,}001$ iterations lifts its ceiling to $78.2$~dB in $4{,}214$~s ($\approx$70~min); ARC-$\phi_1$'s extended run reaches an even higher $133.5$~dB at $2{,}824$ sweeps in $17{,}173$~s ($\approx$4.8~h), Fig.~\ref{fig:native}. At every budget probed here, ARC-$\phi_1$ spends more wall clock per iteration but reaches substantially higher terminal accuracy; the two arms are not run to matched wall clock in this case study (a matched-wall-clock comparison, together with the $\phi_1$ subsolver's small-scale controls and full derivation, is reported in the companion optimizer paper~\cite{podorozhny2026blockarcphi1}, which merges ARC-$\phi_1$ into the blockwise adaptive cubic regularization framework as an interchangeable subsolver).

\paragraph{Two further baselines confirm the mechanism.} We probe two
additional optimizers on the same image-0 saddle-dominated landscape,
same hardware class. L-BFGS (three memory sizes, $m \in \{10, 20,
50\}$, strong-Wolfe line search) does not merely stall like Adam ---
every configuration plateaus at $7.3$~dB (near a constant/DC-only
reconstruction) by iteration $500$ and never moves again through its
own stopping point at iteration $2{,}004$ ($\approx$18~s); memory size
makes no difference. Unlike Adam, a quasi-Newton step has no bounded
escape mechanism for negative curvature, and on this deterministic
full-batch loss there is no gradient noise to dislodge it from the
saddle either --- so L-BFGS never leaves the saddle, rather than
reaching one after productive early progress the way Adam does.
SOAP~\cite{vyas2024soap}, a diagonal (Shampoo/Adam-eigenbasis)
preconditioner tuned by inheriting Adam's own best learning rate
($10^{-4}$) rather than a separate sweep, reaches a best of
$77.65$~dB over the same flat $10^6$-step budget ($5{,}886$~s) ---
close to but not beyond tuned Adam's own $78.2$~dB ceiling at
$4{,}214$~s, and its final iterate ($72.4$~dB) trails its own running
best by more than $5$~dB, evidence of saddle-adjacent instability
rather than a repaired landscape. Both results are consistent with
$\rho = .98$ in Table~\ref{tab:fingerprint}: with the
ill-conditioning this axis-aligned, a diagonal or quasi-diagonal
rescaling of any kind removes the conditioning penalty but does
nothing about the $0.81$ negative spectral mass at init --- only a
step rule with a negative-curvature escape mechanism gets past the
saddle.

Note: The \emph{H/L (hl) ratio} is the ratio of high to low frequency band MSE under
a radial FFT decomposition; values near $1$ indicate uniform per-band
convergence, i.e.\ little spectral bias. It is one of the metrics to evaluate spectral bias. The companion optimizer paper~\cite{podorozhny2026blockarcphi1} reports these ratios on the same FINER fitting task: at convergence, tuned Adam's hl value is $0.22$ while the blockwise cubic-regularized arms are close to $1$ ($1.22$ for the blockwise cubic step, $1.32$ for its Krylov-subspace variant), and at a matched $500$-step budget ARC-$\phi_1$'s residual is already band-uniform (hl $= 1.10$) against $0.81$ for tuned Adam. The same contrast is visible in the extended-budget error maps of Fig.~\ref{fig:native}: the ARC-$\phi_1$ residual is much lower and band-uniform, $\sim$$2\cdot10^{-7}$, while Adam's residual is $\sim$$10^{-4}$ and it is not uniform, most residual power is in DC and low frequency bands. DC is direct-current or zero-frequency component of the image residual’s Fourier spectrum. It is the spatially constant offset, the mean residual across the whole image, so it corresponds to broad background or brightness mismatch rather than fine image detail.

\end{document}